\documentclass{article}

\usepackage{microtype}
\usepackage{graphicx}
\usepackage{subcaption}
\usepackage{booktabs} 
\usepackage{threeparttable}
\usepackage{booktabs}   
\usepackage{graphicx}   
\usepackage{xcolor}

\usepackage{hyperref}
\usepackage{algorithm}
\usepackage{algorithmic}
\usepackage{amsthm}
\usepackage{bm}
\usepackage{graphicx}  

\newtheorem{property}{Property}
\newtheorem{theorem}{Theorem}
\theoremstyle{definition}
\newtheorem{definition}{Definition}  

\usepackage[preprint]{icml2026}

\usepackage{amsmath}
\usepackage{amssymb}
\usepackage{mathtools}
\usepackage{amsthm}

\usepackage[capitalize,noabbrev]{cleveref}

\theoremstyle{plain}

\theoremstyle{definition}

\theoremstyle{remark}

\usepackage[textsize=tiny]{todonotes}

\icmltitlerunning{Submission and Formatting Instructions for ICML 2026}

\begin{document}

\twocolumn[
  \icmltitle{Nested-ReFT: Efficient Reinforcement Learning for Large Language Model Fine-Tuning via Off-Policy Rollouts}



  \icmlsetsymbol{equal}{*}
\begin{icmlauthorlist}
    \icmlauthor{Maxime Heuillet}{ulaval,mila}
    \icmlauthor{Yufei Cui}{huawei}
    \icmlauthor{Boxing Chen}{huawei}
    \icmlauthor{Audrey Durand}{ulaval,mila,cifar}
    \icmlauthor{Prasanna Parthasarathi}{huawei}
\end{icmlauthorlist}
\icmlaffiliation{ulaval}{Université Laval (IID), Canada}
\icmlaffiliation{mila}{Mila - Québec AI Institute, Canada}
\icmlaffiliation{huawei}{Huawei Noah's Ark Lab (Montreal Research Center), Canada}
\icmlaffiliation{cifar}{Canada CIFAR AI Chair}
\icmlcorrespondingauthor{Maxime Heuillet}{maxime.heuillet.1@ulaval.ca}
  \icmlkeywords{Machine Learning, ICML}

  \vskip 0.3in
]



\printAffiliationsAndNotice{}  

\begin{abstract}
Standard Reinforced Fine-Tuning (ReFT) treats off-policyness—the discrepancy between behavior and target policies—as a detrimental artifact to be minimized. We challenge this dogma with Nested-ReFT, a framework that deliberately induces controlled off-policyness during training. By constructing behavior policies as nested instances of the target model, we achieve automatic parameter synchronization. Theoretically, we prove that this formulation yields unbiased gradient estimates under ensemble behavior policies. Empirically, we conduct a systematic study across model scales (1.5B, 7B) and domains (math, code), establishing three key findings: (1) LLMs exhibit "emergent robustness" to structural off-policyness, particularly at larger scales; (2) Mixup Nesting combined with Retrace variance reduction effectively stabilizes the off-policy gap; and (3) Nested-ReFT reduces rollout time by with minimal performance impact. Our work establishes off-policy intensity not as an error to be avoided, but as a tunable design dimension for efficient ReFT.
\end{abstract}

\section{Introduction}

Large language models (LLMs) are increasingly capable at solving complex reasoning problems \citep{cobbe2021trainingverifierssolvemath}. 
This progress is partly driven by LLMs' ability to generate chain-of-thought (CoT) completions, which include the final response to a problem as well as the intermediate reasoning steps helpful to reach that response \cite{wei2022chain}. 
To improve the generalization performance of LLMs, an increasingly popular post-training technique consists of using an LLM to generate CoT completions and use them for fine-tuning \citep{kumar2025llm, shao2024deepseekmath,xie2024monte, silver2016mastering}.

Such post-training techniques are dubbed reinforced fine-tuning (ReFT) and are based on reinforcement learning \citep{sutton2018reinforcement}.
The CoT completions are generated by a behavior LLM policy through a process referred to as rollout. 
Then, the completions are scored using a reward function. 
In domains with verifiable rewards (e.g., math, programming), the reward function is a simple heuristic \citep{luong2024reft, shao2024deepseekmath,liu2025understanding}. 
The scored completions are then used to propagate gradients back to fine-tune the target LLM policy.

The rollout step in ReFT generates a substantial amount of new training data to fine-tune the target LLM through inference.
Appealing generalization gains are achievable using ReFT \citep{luong2024reft}, yet the rollout step comes with a high computational cost. 
While completions are easily verifiable and thus rewards are cheap to acquire, the computational cost of generating completions from a behavior LLM policy can be substantial \citep{kazemnejad2025vinepporefiningcreditassignment, shao2024deepseekmath}. 
This completion generation cost can add up significantly to the compute cost of updating the target LLM's parameters. 

Practitioners have explored several avenues to decrease this cost. 
Open-source frameworks like TRL \citep{vonwerra2020trl} allow doing rollouts with the behavior LLM loaded with VLLM backend \citep{kwon2023efficient}, which specializes in inference acceleration.  
Recently, \citet{zhang2025fastgrpoacceleratingpolicyoptimization} propose using speculative decoding to accelerate rollout inference. 
However, both strategies introduce discrepancies between the behavior and target policy, which can alter training stability.
Indeed, having a distributional gap between the behavior and target LLM policies which corresponds to off-policy learning where the target policy learns from data generated with a different behavior policy \citep{watkins1989learning, sutton2018reinforcement}. 
Off-policy learning introduces fundamental challenges: as the behavior and target policies diverge, importance sampling ratios can exhibit high variance, leading to unstable gradient estimates and potential training collapse \citep{munos2016safeefficientoffpolicyreinforcement, metelli2020importance}. 
This instability has historically motivated practitioners to minimize off-policyness, keeping behavior policies architecturally identical to target policies \citep{luong2024reft, shao2024deepseekmath}.
However, the off-policy gap explored in the ReFT literature to accelerate inference remains either an unintended VLLM software artifact \citep{vonwerra2020trl} or deliberately minimized \citep{yao2025your}, leaving unexplored whether larger off-policy gaps could be viable to decrease the cost of the rollout process.

\paragraph{Research Gap and Questions}
While prior work treats off-policyness in ReFT as an artifact to be minimized, the potential for deliberate off-policy formulations to reduce rollout costs remains unexplored. 
We investigate: (RQ1) Can substantially different behavior policies maintain target model performance? (RQ2) How do off-policyness intensity and patterns affect training stability? (RQ3) Which variance reduction techniques prove most robust? (RQ4) Do patterns generalize across scales and domains?

\paragraph{Contributions}
To investigate these questions, we introduce \texttt{Nested-ReFT}, a framework that deliberately constructs behavior LLM policies as nested instances of the target LLM policy through dynamic nesting. 
The behavior policies are instantiated on-the-fly during training from the target model's current parameters. 
This nesting strategy offers two key advantages: (1) the behavior policy automatically tracks target model improvements throughout training, maintaining relevance without separate synchronization, and (2) layer skipping provides a principled mechanism to tune the degree of off-policyness while preserving architectural compatibility. 
Our contributions are: 
i) We provide the \textbf{first systematic study} of deliberate off-policyness in ReFT to decrease rollout costs, characterizing the stability-efficiency tradeoff across model scales (1.5B, 7B) and domains (math, code). 
ii) We demonstrate theoretically that \texttt{Nested-ReFT} yields unbiased gradient estimates with controlled variance under ensemble behavior policies. 
iii) We empirically validate three variance mitigation strategies (base importance sampling, uncorrected and Retrace-$\lambda$) and three layer skipping patterns (random, ramp-up and mixup) across varying degrees of off-policyness. 

\section{Related works}

\paragraph{Off-Policyness in ReFT} 
Off-policyness in ReFT is multifaceted.
\citet{yao2025your} shows that using VLLM \citep{kwon2023efficient} to generate completions during rollout creates an implementation gap that turns the ReFT to an off-policy learning problem. Their setting assumes off-policyness comes from different backend supporting the behavior and target policy while in our setting the off-policyness comes from using a different parameterized models of different size nested in the target LLM policy. \citet{roux2025tapered} assumes off-policyness emerges from the increasing gap between the target LLM policy and a fixed behavior policy. In contrast, in our setting, the behavior policy is synchronized as it is nested in the parameterization of the target policy (not updated frequently during training). Another source of off-policyness in the widely adopted and open-source framework TRL \citep{vonwerra2020trl} comes from the number of gradient iterations performed on a batch, as the behavior policy generates the batch and then the gap increases as the target policy is being updated at each iterations.

\paragraph{Connection of Nesting with Layer Dropout and Speculative Decoding} 

\citet{Elhoushi_2024} introduced a method to do layer dropout during LLM fine-tuning. However, their method is introduced for supervised fine-tuning while we focus on the different ReFT setting which has distinct training properties (data is self-generated). 
Layer dropout techniques are also leveraged to design self-speculative decoding strategies \citep{xia2024unlocking,Zhang_2024,xia2025swiftontheflyselfspeculativedecoding}, with variants to skip depth-wise or width-wise \citep{narasimhan2025faster}.
These strategies are designed to generate responses at test-time while we are interested in using generated completions from layer skipping to further train a target LLM policy. Additionally, such speculative decoding strategies resort to accept-reject mechanisms to improve completion quality \citep{von195113}, which prevents accessing the probability distribution of the draft model. In our setup, we need to access the probability distribution of both the behavior and target policies to compute the importance sampling ratio, therefore, our proposed layer skipping differs as it omits the accept-reject component typically seen for pure inference applications.
More recently, \citet{zhang2025fastgrpoacceleratingpolicyoptimization} proposes to directly plug a speculative-decoding strategy not based on layer skipping to speed-up the rollout of ReFT. In their setting the behavior policy is a 1-layer transformer that learns to predict the probability distribution of the target LLM policy. Hence, the behavior and target policy belong to distinct algorithmic classes, and both need to be alternatively updated during the training to maintain distributional closeness. In contrast, our nesting strategy instantiates the behavior and target policies under the same class. Additionally, the behavior policy being nested means the parameters are synced throughout training, resolving the need for alternative updates.
Lastly, the proposed method aims to minimize the distributional difference between the behavior and target LLM policies, while our study brings more understanding on the effect of various off-policy patterns and intensity in ReFT.

\section{Problem Setting and ReFT Background}
\label{sec:problem}

\paragraph{Goal:} Given a pretrained LLM, we aim to improve its generalization on reasoning benchmarks through reinforced fine-tuning (ReFT). We measure performance as the average accuracy on multiple held-out benchmarks.

\paragraph{Notation}
Let $\mathcal X$ denote the space of possible problems and $\mathcal{Y}$ denote the space of possible completions.
Given a problem $x_i \in \mathcal{X}$, an LLM policy $\pi_\theta$ defines a conditional probability distribution over completions $\hat{y}_i = (\hat{y}_{i,1}, \dots, \hat{y}_{i,L}) \in \mathcal{Y}$, where $L$ is the number of tokens contained in the completion. 
Let $\hat y_{i,<\ell}$ denote the tokens $(\hat y_{i,1}, \dots, \hat y_{i,\ell-1})$ in a completion $\hat y_i$.
The probability of sampling completion $\hat y_i$ given a problem $x_i$ is defined in an auto-regressive manner:
$$
\pi_\theta(\hat y_i|x_i)=\Pi_{\ell=1}^L \pi_\theta(\hat y_{i, \ell}|x_i, \hat y_{i, <\ell}),
$$
where $\pi_\theta(\hat{y}_{i,\ell} | x_i, \hat{y}_{i,<\ell})$ is the probability of sampling token $\hat{y}_{i,\ell}$ given the problem $x_i$ and the previous tokens $\hat{y}_{i,<\ell}$.

When applying LLMs to verifiable rewards domains such as math reasoning or programming, it is useful to distinguish chain-of-thought (CoT) completions $\mathcal{Y}^{\text{cot}} \subset \mathcal Y$ from their value answers $\mathcal{Y}^{\text{val}} \subseteq \mathcal{Y}^{\text{cot}}$. 
The value is the exact solution to a problem (math answer, generated function), while a CoT includes both the reasoning steps and the value. 
We assume access to a deterministic extraction function $v: \mathcal{Y}^{\text{cot}} \mapsto \mathcal{Y}^{\text{val}}$ that extracts value answers from CoT completions.

\subsection{Reinforced fine-tuning (ReFT)}

Let $\theta_{\text{ref}}$ denote the parametrization of a pretrained LLM policy.
Reinforced Fine-Tuning (ReFT) aims to further train $\pi_{\theta_{\text{ref}}}$ by leveraging reward signal on a given dataset $D$. 
The dataset $D = {(x_i, y_i^{\text{cot}})}_{i=1}^{|D|}$ contains problems $x_i \in \mathcal{X}$ and their associated CoT solutions $y_i^{\text{cot}} \in \mathcal{Y}^{\text{cot}}$. 
Note that none of these problems are contained in the testing benchmarks.

\paragraph{Training procedure with GRPO.} 
We focus on performing ReFT with the reinforcement learning algorithm GRPO \citep{shao2024deepseekmath}. 
ReFT proceeds for $E_{\text{rft}}$ epochs over dataset $D$. Each epoch processes the dataset in batches of $B$ problems. For each batch:
1)  \textit{Behavior Policy Rollout:} The behavior policy $\eta_\chi$ generates $G$ completions per problem, yielding $B \times G$ total completions. Each completion is scored by a reward function that checks answer correctness. 2) \textit{Target Policy Update:} We perform $K$ gradient update iterations on the scored completions. Within each iteration, completions are processed in mini-batches of size $B_{\text{mini}}$, with gradients accumulated over $B \times G / B_{\text{mini}}$ steps before each parameter update.
This yields $S = \frac{|D|}{B} \cdot K \cdot E_{\text{rft}}$ total gradient steps. 

\paragraph{Importance sampling} 
In off-policy RL, importance sampling accounts for the distributional gap between the behavior and the target policies.
More specifically, the $\texttt{base}$ importance sampling ratio reweights the rewards with:
\begin{equation*}
 h_{\texttt{base}}(\hat y, x; \pi_\theta, \eta_\chi ) = \frac{\pi_\theta(\hat{y} \mid x)}{\eta_\chi(\hat{y} \mid x)} = 
\prod_{\ell=1}^L \frac{\pi_\theta(\hat{y}_\ell \mid x, \hat{y}_{<\ell})}{\eta_\chi(\hat{y}_\ell \mid x, \hat{y}_{<\ell})}.
\end{equation*}
The importance sampling ratio can suffer high variance, especially when the behavior and target policies diverge significantly \citep{xie2019towards}, which can negatively affect training quality. In classical RL, variance reduction techniques were proposed to stabilize training \citep{munos2016safeefficientoffpolicyreinforcement,metelli2020importance} and their applicability to ReFT and GRPO under various off-policyness intensity and patterns remains an open question.

\paragraph{Training setup}

In this study, we focus on the single iteration case ($K=1$). This means at a specific step $s$, we have $\eta_\chi=\pi_{\theta_{\text{s-1}}}$, where $\theta_{\text{s-1}}$ corresponds to the target policy parametrization from the previous gradient step.
In the baseline ReFT case (i.e. without Nested-ReFT), there is no architectural difference between $\pi_\theta$ and $\pi_{\theta_{\text{s-1}} }$, so the importance sampling ratio is $\textbf{1}$. The case with $K>1$ introduces other form of off-policyness that would interact with the research questions brought by Nested-ReFT. We explore the possibility of generating completions from a behavior policy that architecturally differs but is nested from the target policy, thus causing a more substantial increase in the off-policyness of the learning problem.

\section{ The Nested-ReFT framework }

In this Section, we introduce the \texttt{Nested-ReFT} framework. Nested-ReFT introduces two orthogonal design axes: (i) how off-policy behavior policies are instantiated (nesting patterns), and (ii) how the resulting distributional shift is corrected (variance mitigation).
\textbf{Importantly, our goal is to navigate the performance-stability trade-off optimally, relative to the ReFT baseline.} We hypothesize that the standard ReFT approach, where behavior and target policies are identical, serves as a performance ceiling.

We propose different design patterns for nesting, and also identify solutions to decrease the variance of the importance sampling ratio. Algorithm \ref{alg:nest_reft} summarizes the framework, with \textcolor{purple}{purple} highlighting key novelty compared to current ReFT. A notable benefit is that the framework can be seamlessly plugged into existing ReFT frameworks. Though we experiment the framework with the popular GRPO algorithm \citep{shao2024deepseekmath}, \texttt{Nested-ReFT} is agnostic and can be combined with any sampling based RL post-training technique \citep{ahmadian2024basicsrevisitingreinforcestyle,ziegler2020finetuninglanguagemodelshuman}.

Consider the target model $\pi_\theta$ to fine-tune and a behavior model $\eta$. 
We instantiate nested models with layer skipping, which consists of selecting a set of layers that are not used (skipped) during the forward pass, acting like a shortcut. 

\begin{definition}[Transformer layer indices]
Let $T_{\eta} = \{0, 1, \dots, N\}$ denote the set of transformer layer indices in model $\eta$, where $N+1 = |T_{\eta}|$ is the total number of layers.
\end{definition}

Following prior work \citep{Elhoushi_2024, fan2019reducing}, we exclude the first two and last two layers from skipping, as boundary layers are critical for representation \citep{van_Aken_2019}. This yields an eligible set of layers $V_{\eta} = \{2, 3, \dots, N-2\}$ with $|V_{\eta}| = N - 3$.

Given a user-specified skip ratio $x \in [0,1]$, the number of layers to skip is $U_x = \lceil |T_\eta| \cdot x \rceil$.
We compute $U_x$ from the total layer count $|T_\eta|$ (rather than from $|V_{\eta}|$) to keep the skip ratio interpretable and independent of boundary protection.

\begin{definition}[Nesting function]
The nesting function $f_{x}: \eta \times \mathbb{N} \rightarrow \{0,1\}^{|T_\eta|}$ outputs a binary mask $\sigma = f_{x}(\eta, s)$ at step $s$, where $\sigma_i = 1$ indicates layer $i$ should be skipped. The mask satisfies: 1)  Correct skip count: $\sum_{i \in T_\eta} \sigma_i = U_x$, 2) Boundary protection: $\sigma_i = 0$ for $i \in \{0, 1, N-1, N\}$.
\end{definition}

At gradient step $s$, in the baseline ReFT framework, the behavior model is defined as $\eta_{s} = \pi_{\theta_{s-1}}$ (the target model from the previous step). In Nested-ReFT, we use instead the nested behavior model $\eta'_{s}$ instantiated by applying $f_{x}(\eta_s, s)$ to deactivate the masked layers during the forward pass.

\begin{figure}
    \centering
    \includegraphics[width=0.99\linewidth]{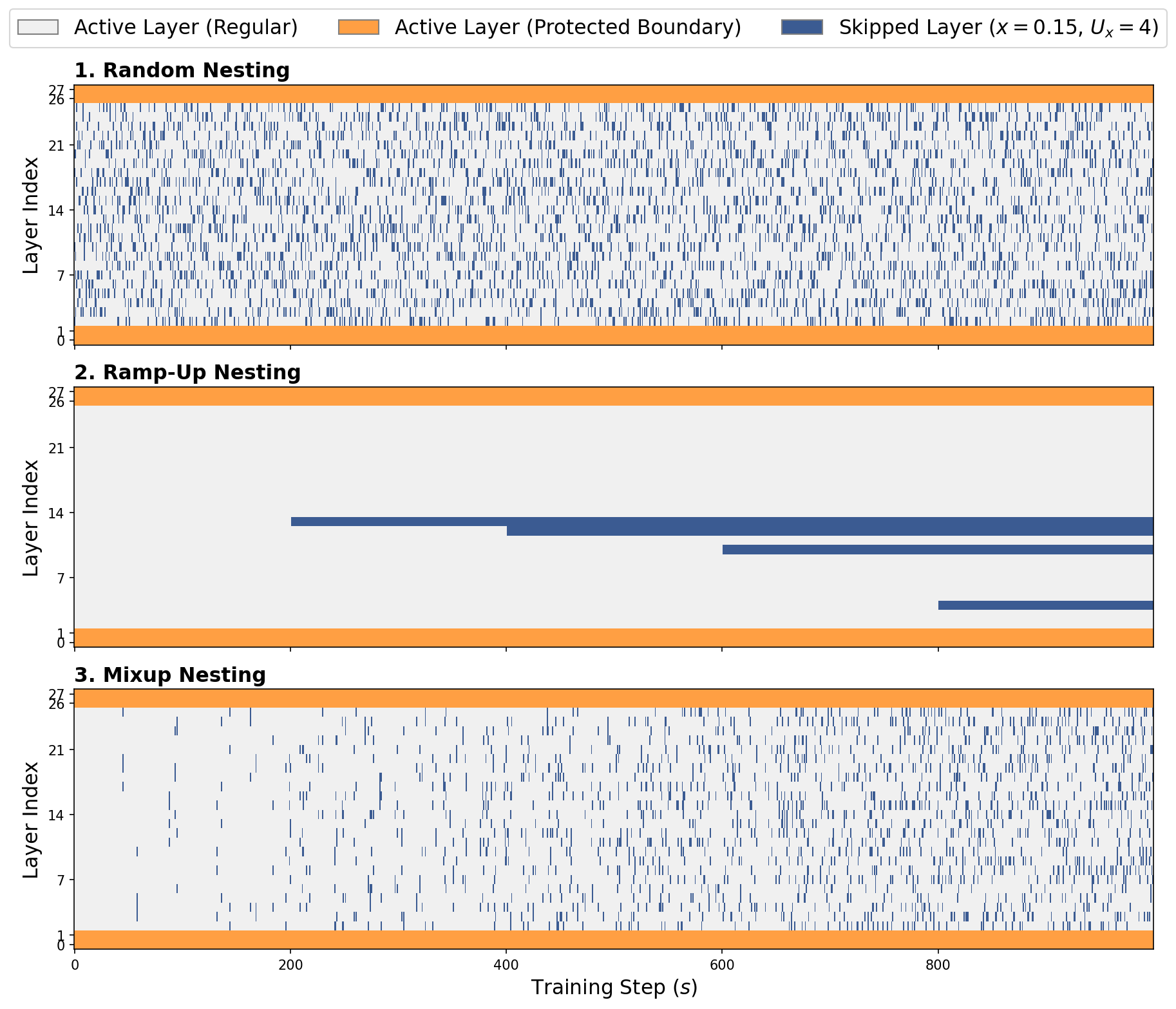}
    \caption{Visualization of three nesting strategies across training.}
    \label{fig:schedules}
\end{figure}

\subsection{Nesting Patterns}

Since \texttt{Nested-ReFT} instantiates a nested behavior LLM policy at each step $s \leq S$, we obtain throughout training an ensemble of nested models $\mathcal{Z}_f = \{\eta'_s\}_{s=1}^S$.
We explore three strategies examplified in Figure \ref{fig:schedules} for constructing this ensemble, each offering different trade-offs between exploration diversity, training stability, and off-policy intensity.

\paragraph{Random Nesting}
At each step $s$, the function $f_{x}^{\text{rand}}(\eta_s, s)$ samples $U_x$ layers uniformly from $V_{\eta}$ using a step-dependent random seed. 
This produces maximum diversity in the ensemble $\mathcal{Z}_{f^{\text{rand}}}$, with each nested model skipping different layer combinations.
This strategy draws inspiration from dropout regularization \citep{srivastava2014dropout}, which improves generalization by training an implicit ensemble of sub-networks. 
Similarly, random nesting exposes the target policy to corrections from a diverse set of behavior policies, potentially improving robustness to architectural perturbations.
However, the constant off-policy gap from step one may introduce high variance in the importance sampling ratio, particularly in early training when the policy is rapidly changing. The strategy closely relates to \citet{Elhoushi_2024} who proposed layer dropout during supervised fine-tuning and that we generalize for ReFT.

\paragraph{Ramp-Up Nesting}
The function $f_{x}^{\text{ramp}}(\eta_s, s)$ pre-selects at initialization a fixed candidate set $C \subset V_{\eta}$ with $|C| = U_x$ layers. Then, the number of active skips from $C$ increases gradually over training: 1) Warm-up phase ($s < w \cdot S$ for warmup ratio $w$): No layers are skipped ($\sigma = \mathbf{0}$), allowing the policy to stabilize under on-policy learning, 2) Ramp-up phase ($s \geq w \cdot S$): The number of skipped layers grows linearly from $0$ to $U_x$. At step $s$, the first $n_s$ layers from $C$ are skipped, where $n_s = \min\left(\left\lceil U_x \cdot \frac{s - wS}{S - wS} \right\rceil, U_x\right)$. This design follows the principle of curriculum learning \citep{bengio2009curriculum} where the model first learns under easier on-policy conditions before facing gradually harder off-policy settings.
The deterministic mask ensures that the same layers are consistently skipped, reducing diversity in $\mathcal{Z}_{f^{\text{ramp}}}$ but allowing the model to adapt to a predictable nesting pattern.
However, the discontinuous transition from on-policy to aoff-policy and the discrete off-policy ramp-up may cause optimization instability.

\paragraph{Mixup Nesting}
The function $f_{x}^{\text{mix}}(\eta_s, s)$ varies both \textit{which} layers are skipped and the \textit{proportion} of completions using skipping within each step. At step $s$, a fraction $p_s \in [0, m]$ of rollout completions are generated with layer skipping, while the remaining $(1 - p_s)$ completions use the full model (we set $m = 0.9$ to preserves at least $10\%$ on-policy completions). The proportion $p_s$ increases linearly over $S$ steps.
This strategy combines ideas from mixup regularization \citep{zhang2018mixupempiricalriskminimization} and relates to data centric off-policy correction such as replay buffers \citep{lin1992self}. 
By maintaining a mixture of on-policy and off-policy samples throughout training, mixup nesting provides a smoother transition than annealed nesting while offering more stability than pure random nesting.
The preservation of clean samples acts as an anchor, reducing variance in gradient estimates and preventing the optimization from drifting due to compounding off-policy errors.
The stochastic mask selection, combined with gradual proportion increase, yields a curriculum over both the \textit{intensity} (proportion of nested completions) and \textit{diversity} (which layers are skipped).

\begin{algorithm}
\caption{ \texttt{Nested-ReFT}}
\begin{algorithmic}[1]
\REQUIRE Target model $\pi_\theta$ (LLM), dataset $\mathcal{D}$, skip ratio $x \in (0, 1)$, SFT epochs $E_{\text{sft}}$, RFT gradient steps $S$, choice of $z \in \{\texttt{rand}, \texttt{mix}, \texttt{ramp} \}$ and $m \in \{ \texttt{base}, 1, \lambda \}$

\vspace{1mm}
\STATE \textbf{Step 1: Supervised Fine-Tuning (SFT)}
\FOR{$e = \{1, \dots, E_{\text{sft}} \} $}
    \STATE Train $\pi_\theta$ on $(x, y^{cot}) \sim \mathcal{D}$ using cross-entropy loss
\ENDFOR

\vspace{1mm}
\STATE \textbf{Step 2: Reinforced Fine-Tuning (ReFT)}

\FOR{$s = \{ 1, \dots, S \}$}
        \STATE Sample batch of prompts
        \STATE \textcolor{purple}{Set $\eta_s = \pi_{\theta_{s-1}}$} 
        \STATE \textcolor{purple}{Sample skip set with $f^z_{x}(\eta_s,s)$ } 
        \STATE \textcolor{purple}{Deactivate layers in $\eta_s$ using $f^z_{x}(\eta_s,s)$ to get $\eta'_{s}$}
        \STATE \textcolor{purple}{Generate $G$ samples for each $x$ in batch using $\eta'_s$}
        \STATE Score samples using reward model
        \STATE \textcolor{purple}{Compute stabilization $h_m(\eta', \pi_{\theta})$}
        \STATE \textcolor{purple}{Update $\pi_{\theta_s}$ with rewards and $h_m(\eta'_s, \pi_{\theta_s})$}
        \STATE Set $\pi_{\theta_{s-1}}=\pi_{\theta_s}$  
\ENDFOR
\STATE Return $\pi_{\theta_S}$
\end{algorithmic}
\label{alg:nest_reft}
\end{algorithm}

\subsection{Variance Mitigation Strategies}
\label{subsec:stabilizing}

The nesting patterns increase the distributional gap between the behavior and target policies, potentially causing high variance in the importance sampling ratio. We explore three formulations of the importance weight function $h_m(\cdot; \cdot)$, where $m \in \{\texttt{base}, \texttt{unc}, \lambda\}$.

\paragraph{Base importance sampling} The function corresponds to the standard importance sampling ratio, defined in Section \ref{sec:problem}. This provides unbiased gradient estimates but can exhibit high variance when the behavior and target policies diverge significantly.

\paragraph{Uncorrected off-policy} 
The function ignores the distributional mismatch entirely, by being $h_{\texttt{unc}}(\hat y, x; \pi_{\theta_s}, \eta'_s) = \mathbf{1}$. This choice is motivated by three considerations:
(1) \textit{Coupling with nesting}: Since our behavior policy $\eta'_s$ is instantiated from $\pi_{\theta_{s-1}}$ via layer skipping and shares most parameters with the target policy, the distributional shift may be sufficiently small that correction provides minimal benefit relative to the variance cost.
(2) \textit{Variance reduction}: While setting the ratio to 1 introduces bias, it eliminates the multiplicative variance from importance weights, which can dominate under substantial off-policyness \citep{metelli2020importance}.
(3) \textit{Empirical precedent}: Successful RL implementations in different settings use unit weights or omit importance sampling corrections entirely \citep{vonwerra2022trl,yao2025your}, suggesting the bias-variance tradeoff may favor reduced variance in some high-dimensional settings.

\paragraph{Retrace-$\lambda$} 
The function truncates and scales the importance ratio such that $h_{\lambda}(\hat y, x; \pi_{\theta_s}, \eta'_s) = \lambda \min\bigl(1, h_{\texttt{base}}(q, a; \pi_{\theta_s}, \eta'_s)\bigr)$. This approach follows \citet{munos2016safeefficientoffpolicyreinforcement}, who demonstrated that clipping importance weights stabilizes training under varying degrees of off-policyness in traditional RL. 
While Retrace-$\lambda$ was originally designed for the class of multi-step TD control with Q-functions, we investigate its applicability to the class of policy gradient methods to which GRPO belongs, hypothesizing that the truncation mechanism transfers to the GRPO objective.

\section{Analysis of Nested-ReFT}

Consider a behavior model $\eta$ that contains $|T_\eta|$ identical transformer layers. Each transformer layer has a computational complexity:
$$ \mathcal{C}_{\text{layer}} = O(L^2 d + L d^2),$$
where $L$ is the generated sequence length and $d$ is the hidden dimension \citep{vaswani2017attention}.

\begin{property}[Complexity with Nested-ReFT]
Given a model $\eta$ with $|T_\eta|$ transformer layers, skipping $U_x$ layers yields a nested model $\eta'$ with computational complexity:
$$ \mathcal{C}_{\eta'} = O\bigl((|T_\eta| - U_x) \times \mathcal{C}_{\text{layer}}\bigr).$$
\end{property}

The complexity of rollout inference of $\eta'$ is reduced proportionally to the number of skipped layers $U_x$ through the skip ratio $x\%$. Assuming fixed generation length $L$ and fixed hidden dimension $d$, the layer skipping achieves a linear complexity improvement.

\subsection{Off-Policy Gradient Estimation with Nested Ensembles}

We analyze the bias and variance properties of gradient estimates under \texttt{Nested-ReFT}. Since the behavior policy $\eta'$ differs from the target policy $\pi_\theta$, we operate in the off-policy regime and must account for the distributional mismatch.

\paragraph{Setup.} 
Let $s_\ell = (x, \hat{y}_{<\ell})$ denote the state at token position $\ell$, comprising the prompt $x$ and previously generated tokens. The action is the next token $\hat{y}_\ell$. We assume access to an advantage function $A^\pi(s_\ell, \hat{y}_\ell)$ estimated under the target policy $\pi$ (e.g., via GRPO's group-relative rewards).

\paragraph{Off-Policy Policy Gradient.}
The standard off-policy policy gradient estimator corrects for the distributional mismatch between behavior and target policies using importance sampling:
\begin{equation}
\mathbb{E}_{\hat{y} \sim \eta'(\cdot|x)} \left[ h_{\texttt{base}}(\hat{y}, x; \pi_\theta, \eta') \cdot \nabla_\theta \log \pi_\theta(\hat{y} | x) \cdot A^{\pi}(x, \hat{y}) \right],
\end{equation}
where $A^{\pi}(x, \hat{y})$ is the advantage under the target policy $\pi_\theta$.

\paragraph{Ensemble of Nested Behavior Policies.} In \texttt{Nested-ReFT}, we use an ensemble of nested behavior policies $\mathcal{Z} = \{\eta'_s\}_{s=1}^{S}$ throughout training, where each $\eta'_s$ is instantiated via nesting from the current target parameters. Let $q(\sigma)$ denote the distribution over layer masks $\sigma$ induced by the nesting pattern $z \in \{\texttt{rand}, \texttt{ramp}, \texttt{mix}\}$. This defines a mixture behavior policy:
\begin{equation}
\bar{\eta}_{q}(\hat{y}_\ell | s_\ell) = \mathbb{E}_{\sigma \sim q}[\eta_\sigma(\hat{y}_\ell | s_\ell)],
\end{equation}
where $\eta_\sigma$ is the nested policy with mask $\sigma$.

\begin{theorem}[Unbiased Gradient Estimation with Ensemble Behavior Policies]
\label{thm:unbiased}
Let the behavior policy ensemble $\mathcal{Z}$ satisfy absolute continuity: for all $\eta_\sigma \in \mathcal{Z}$, $\eta_\sigma(\hat{y}|x) > 0$ whenever $\pi_\theta(\hat{y}|x) > 0$. Then, when using $h_{\texttt{base}}$, the expected policy gradient estimator over the ensemble is unbiased:
\begin{align}
&\mathbb{E}_{\sigma \sim q}\Big[\mathbb{E}_{\hat{y} \sim \eta_\sigma}\big[h_{\texttt{base}}(\hat{y}, x; \pi_\theta, \eta_\sigma) \nonumber \\
&\quad \cdot \nabla_\theta \log \pi_\theta(\hat{y}|x) \cdot A^{\pi}(x, \hat{y})\big]\Big] = \nabla_\theta J(\pi_\theta).
\end{align}
\end{theorem}

\begin{proof}
For any fixed mask $\sigma$, by the importance sampling identity:
\begin{align}
&\mathbb{E}_{\hat{y} \sim \eta_\sigma}\left[h_{\texttt{base}}(\hat{y}, x; \pi_\theta, \eta_\sigma) \cdot \nabla_\theta \log \pi_\theta(\hat{y}|x) \cdot A^{\pi}(x, \hat{y})\right] \\
&= \mathbb{E}_{\hat{y} \sim \eta_\sigma}\left[\frac{\pi_\theta(\hat{y}|x)}{\eta_\sigma(\hat{y}|x)} \cdot \nabla_\theta \log \pi_\theta(\hat{y}|x) \cdot A^{\pi}(x, \hat{y})\right] \\
&= \sum_{\hat{y}} \eta_\sigma(\hat{y}|x) \cdot \frac{\pi_\theta(\hat{y}|x)}{\eta_\sigma(\hat{y}|x)} \cdot \nabla_\theta \log \pi_\theta(\hat{y}|x) \cdot A^{\pi}(x, \hat{y}) \\
&= \sum_{\hat{y}} \pi_\theta(\hat{y}|x) \cdot \nabla_\theta \log \pi_\theta(\hat{y}|x) \cdot A^{\pi}(x, \hat{y}) \\
&= \mathbb{E}_{\hat{y} \sim \pi_\theta}\left[\nabla_\theta \log \pi_\theta(\hat{y}|x) \cdot A^{\pi}(x, \hat{y})\right] = \nabla_\theta J(\pi_\theta).
\end{align}
Since this holds for every $\sigma$, taking expectations over $\sigma \sim q$ preserves the equality. The theoretical guarantee holds in expectation over training steps.
\end{proof}

\paragraph{Variance Considerations.}
While $h_{\texttt{base}}$ yields unbiased gradients, the importance ratio can exhibit high variance when behavior and target policies diverge, as the product of per-token ratios can grow with sequence length \citep{metelli2020importance}.  Following V-trace \citep{espeholt2018impalascalabledistributeddeeprl} and tapered importance weighting \citep{roux2025tapered}, we consider truncated estimators introduced in Section~\ref{subsec:stabilizing}:
\textbf{Uncorrected ($h_1$):} Setting $h_1 \equiv 1$ eliminates variance from importance weights but introduces bias, as we no longer correct for the distributional mismatch. This is equivalent to treating off-policy samples as if they were on-policy.
\textbf{Retrace-$\lambda$ ($h_\lambda$):} The truncated ratio $h_\lambda = \lambda \cdot \min(1, h_{\texttt{base}})$ bounds the maximum weight, reducing variance while introducing controlled bias. The parameter $\lambda \in (0, 1]$ trades off bias against variance: smaller $\lambda$ reduces variance but increases bias.

Crucially, the nested structure of \texttt{Nested-ReFT} provides implicit variance control: since $\eta'_s$ shares most parameters with $\pi_{\theta_{s-1}}$ and differs only in skipped layers, the distributional gap remains bounded by the architectural similarity, making the bias-variance trade-off more favorable than with arbitrary behavior policies.

\section{Experimental setup}

We focus on two distinct verifiable rewards domains: math reasoning and programming. We consider two different datasets for fine-tuning, namely GSM8k \citep{cobbe2021trainingverifierssolvemath}, and MDPP (train set) \citep{austin2021program}.
We fine-tune four distinct LLMs, namely \textit{Qwen2.5-Math-Instruct} and \textit{Qwen2.5-Coder-Instruct} \citep{yang2024qwen25mathtechnicalreportmathematical} of sizes $1.5$B and $7$B (see Table \ref{tab:skipped_layers} in the Appendix).
To measure generalization performance we use held-out benchmarks. For math reasoning, we use AIME2024 \citep{li2024numinamath}, AMC \citep{li2024numinamath}, MATH500 \citep{hendrycks2021measuringmathematicalproblemsolving}, Minerva \citep{lewkowycz2022solving}, and Olympiad \citep{he2024olympiadbenchchallengingbenchmarkpromoting}. For programming, we use the test sets of the MDPP \citep{austin2021program} and of the Human-Eval datasets \citep{chen2021evaluating}. All the training and evaluation hyper-parameters are reported in the Appendix. 

\section{Empirical Results}

\subsection{Influence of the nesting pattern}
\label{sec:nesting_pattern_results}
In this experiment, we compare the influence of different nesting patterns on the learning quality using a 7B Math Instruct Model, with base variance mitigation at skipping ratios $5\%$ and $15\%$. The nesting pattern $f^{\texttt{rand}}$ is a baseline based on LayerDrop from \citet{fan2019reducing}, proposed originally for supervised fine-tuning (generalized to ReFT).

The ramp-up pattern allows the model to learn well in the first stage (warm-up set to 200 steps) but then the reward sharply degrades and the clipping of the importance sampling ratio sharply increases possibly causing instability. Furthermore, the gradient norm hits high values (above 10 and 100) which are captured through gradient clipping, but also indicate learning instability. The warm-up phase of the ramp-up pattern first lets the learning be pure on-policy (behavior and target LLM policies are identical) but the introduction of skipping after the warm-up converts the problem to off-policy mid-training which makes the problem significantly harder. 
In contrast, the gradient norms of the random baseline are less volatile and the clipping of the importance ratio remains steady throughout training. However, the random baseline shows no reward progression over $1000$ steps of training. 
The mixup pattern demonstrates reward progression when the skipping is light (i.e. $5\%$) and slow decay with heavy skipping (i.e. $15\%$). The clipping of the importance sampling ratio steadily increases while the gradient norm remains the lowest, indicating stability. 

A key distinction between patterns lies in the performance-stability trade-off. While random and ramp-up yield faster wall-clock training times, mixup provides the most favorable balance: it maintains stable gradient norms throughout training, avoids the sharp instabilities observed with ramp-up, and unlike the random baseline, demonstrates consistent reward progression. This makes mixup the preferred nesting pattern when training stability is prioritized alongside final performance, even if it offers less runtime acceleration compared to the other patterns.

\begin{figure}[htbp]
    \centering
    \includegraphics[width=0.99\linewidth]{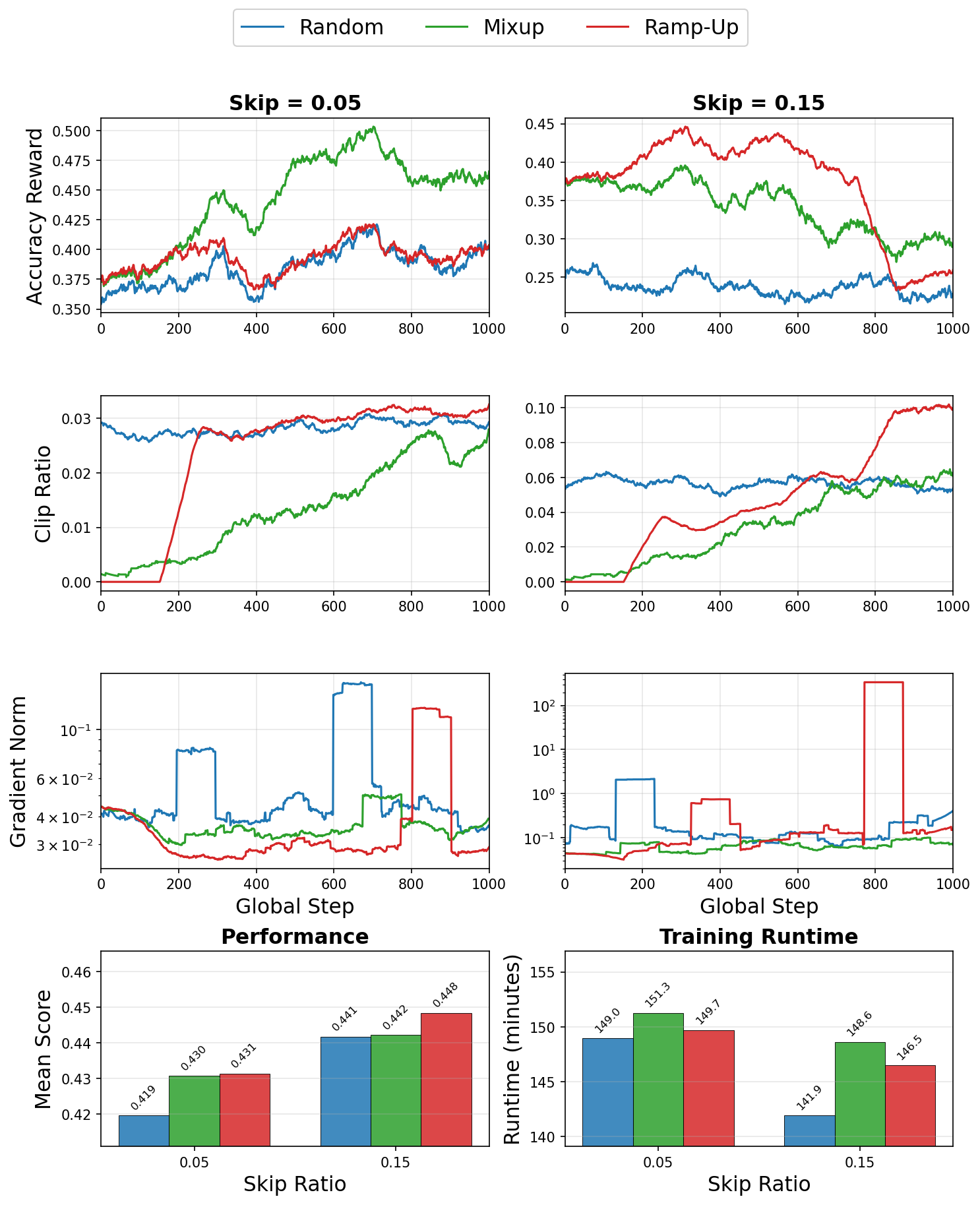}
    \caption{Comparison of different patterns for nesting across different training stability metrics. Smoothing to 100 steps.}
    \label{fig:placeholder}
\end{figure}

\subsection{Impact of off-policy roll-outs on performance} 

In this Section, for statistical significance, we run 3 random seeds for each measurement. We report performance and runtime variations to the baseline ReFT configuration (diplayed as a black star). This baseline consists in fine-tuning at ratio $x=0\%$ and with variance mitigation method $m=\texttt{base}$, which corresponds to classic GRPO without off-policy acceleration from nesting.

For the Nested-ReFT instances, we consider a proportion of skipped layers $x \in \{5\%, 10\%, 15\%\}$ motivated by the training stability results from Section \ref{sec:nesting_pattern_results}. Similar cap ranges are used in related works \citep{fan2019reducing}. All the instances use the mixup nesting pattern which training properties were validated in the previous section. We consider variance mitigation strategies $h_m$, with $m\in\{\texttt{base}, \texttt{uncorrected}, \lambda\}$. For Retrace-$\lambda$, we consider the instances in $\lambda \in \{ 0.1, 1\}$ where $\lambda$ controls the bias-variance trade-off ($\lambda = 0.1$ aggressively downweights off-policy samples for stability, while $\lambda = 1$ applies only the clipping).

\begin{figure}[htbp]
    \centering
    \includegraphics[width=\linewidth]{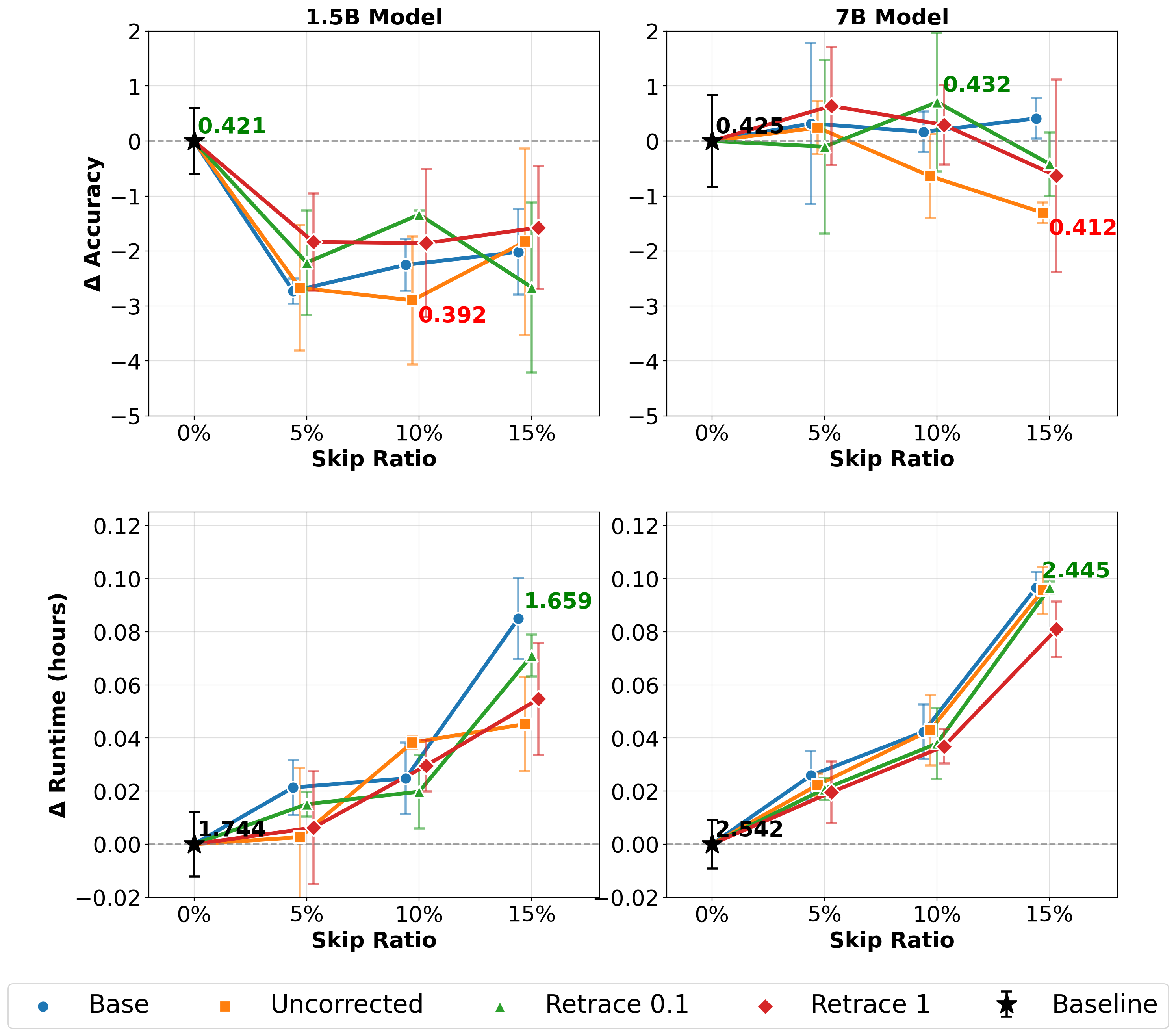}
    \caption{Fine-tuning on GSM8k. \textcolor{red}{Red} annotations indicate the smallest value, and \textcolor[HTML]{2ca02c}{Green} annotations the largest value.}
    \label{fig:gsm8k}
\end{figure}

\paragraph{Influence of nesting on generalization performance}

\begin{figure}[t]
  \centering
  \begin{subfigure}[b]{0.48\columnwidth}
    \centering
    \includegraphics[width=\textwidth]{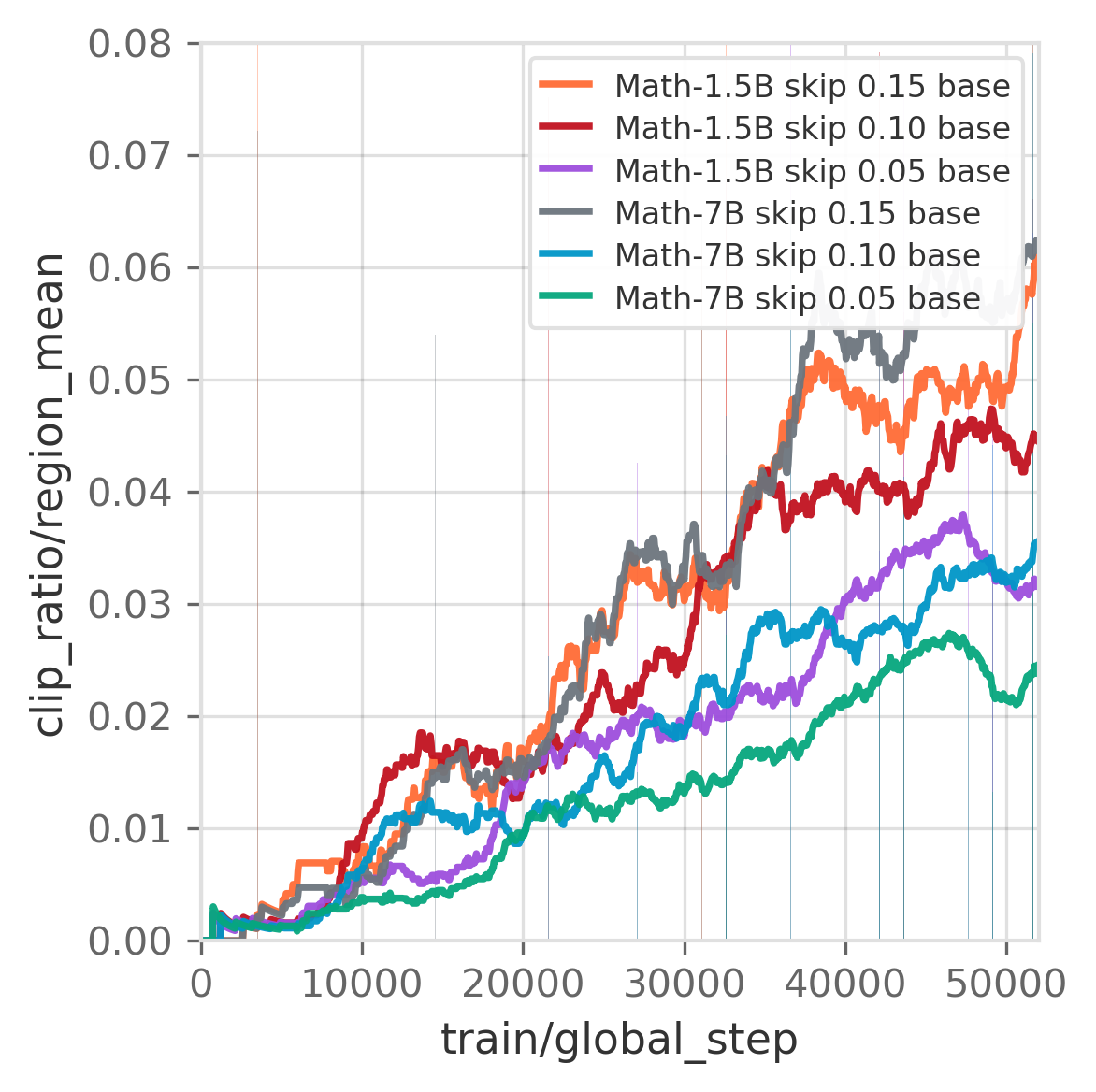}
    \caption{Math (GSM8k)}
    \label{fig:final2}
  \end{subfigure}
  \hfill
  \begin{subfigure}[b]{0.48\columnwidth}
    \centering
    \includegraphics[width=\textwidth]{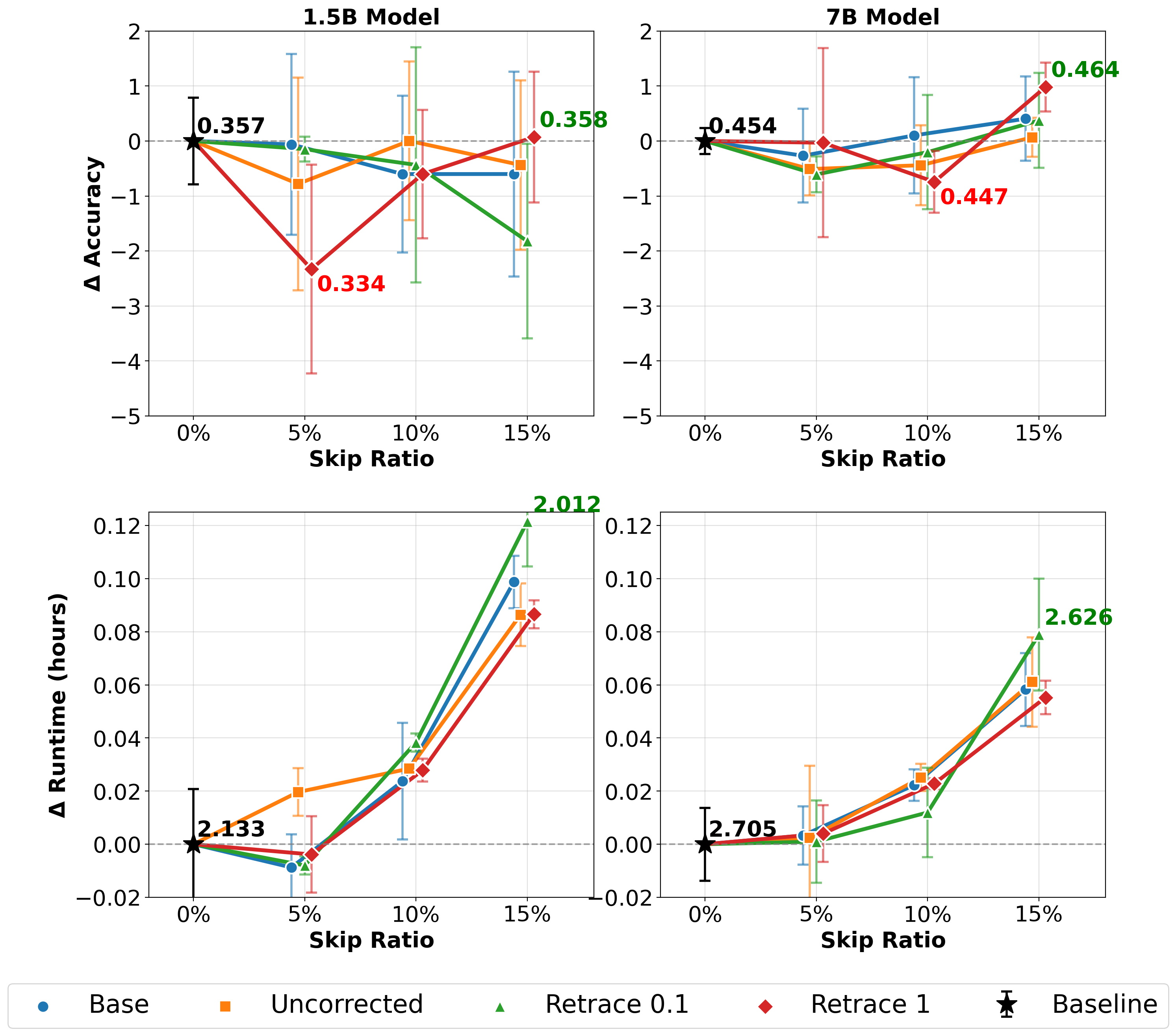}
    \caption{Programming (MBPP)}
    \label{fig:final4}
  \end{subfigure}
  \caption{Running averages (window=100) of the clip rate of importance sampling (base).}
  \label{fig:importance_sampling}
\end{figure}

\begin{figure}[htbp]
    \centering
    \begin{minipage}{\linewidth}
        \centering
        \includegraphics[width=\linewidth]{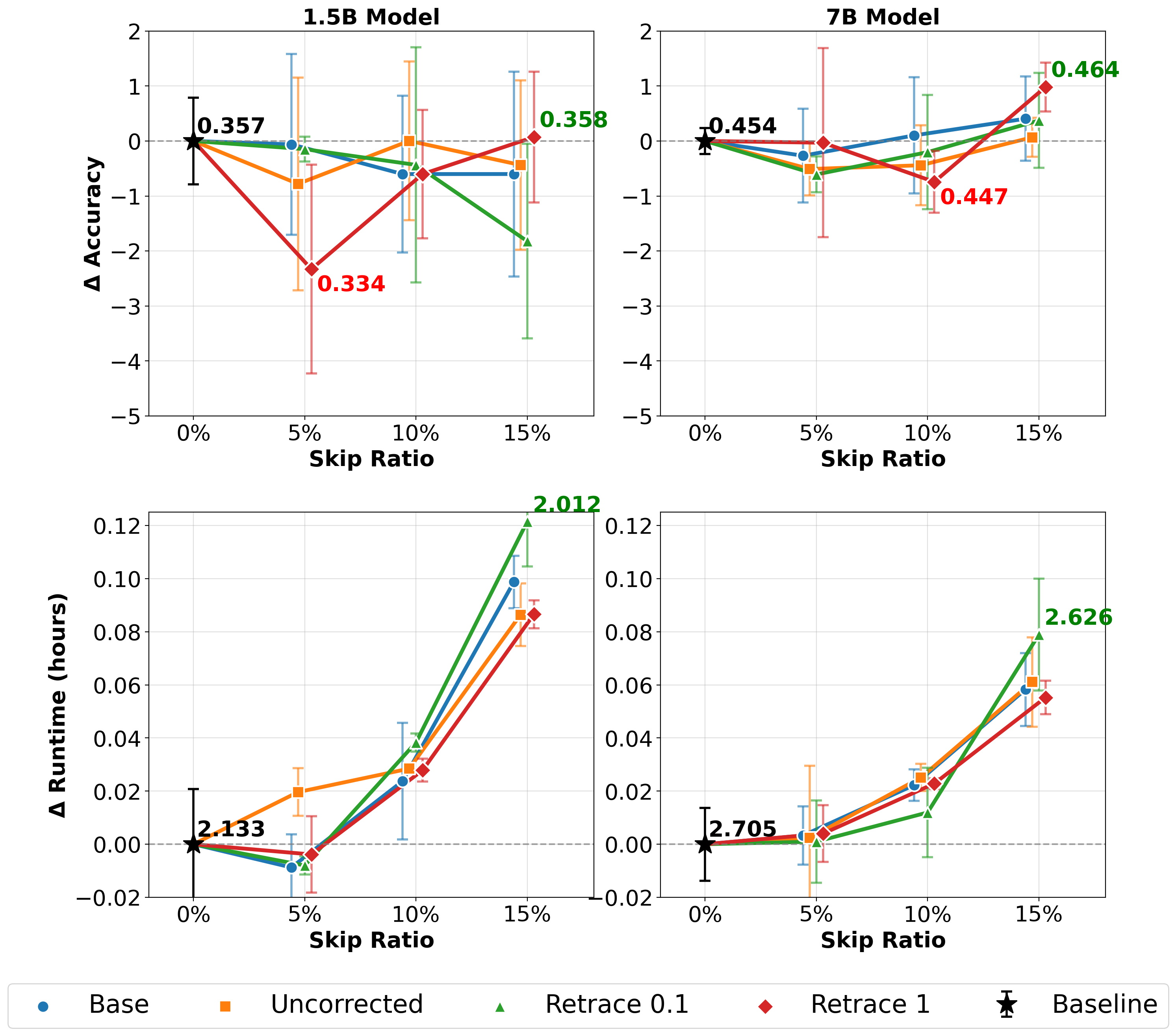}
    \end{minipage}
    \caption{Fine-tuning on MBPP. \textcolor{red}{Red} annotations indicate the smallest value, and \textcolor[HTML]{2ca02c}{Green} annotations the largest value.}
    \label{fig:mbpp}
\end{figure}

The results are displayed in Figure \ref{fig:gsm8k} for math reasoning and \ref{fig:mbpp} for programming.
Performance variations to the baseline must be contextualized with the compute-performance tradeoff inherent to Nested-ReFT. Negative performance variations are expected as completions are generated from smaller models. The goal is to obtain the smallest performance drops to the baseline (black star). We can observe on the figure that the performance obtained with Nested-ReFT instances is often included withing the performance range of the baseline measured across the three seeds.
From Table \ref{tab:variation_from_baseline}, we observe that the worst case performance gap to the baseline widens more with nesting on smaller models (1.5B) compared to bigger models (7B), and this trend is maintained across math reasoning and programming domains. The trend is further supported in Figure \ref{fig:importance_sampling}, where the proportion of clipped values in the importance sampling, that is smaller for 7B models compared to 1.5B models.
These results suggest that the residual connections in larger LLMs may be more robust than in smaller scale LLMs, as the residual connections are highly solicited in Nested-ReFT due to the use of layer skipping \citep{vaswani2017attention, he2016deep}
Importantly, we highlight that some instances of \texttt{Nested-ReFT} yield performance improvements over the baseline, while involving the generation of samples on a smaller model, indicating that nested models can deliver similar or better effect as full models.
These results showcasing minor performance gaps around the baseline corroborate the hypothesis that off-policy generations using \texttt{Nested-ReFT} have limited influence on the generalization performance.

\begin{table}[htbp]
\centering
\caption{Best and Worst Case Performance Variation from Baseline (skip ratio of 0\%). Values in perf. points.}
\label{tab:variation_from_baseline}
\begin{tabular}{llcc}
\toprule
\textbf{Domain} & \textbf{Model} & \textbf{Worst} & \textbf{Best} \\
\midrule
coding & Coder-1.5B-Instruct & -3.86 & +1.83 \\
coding & Coder-7B-Instruct & -1.62 & +1.62 \\
math & Math-1.5B-Instruct & -4.93 & -0.17 \\
math & Math-7B-Instruct & -2.61 & +2.86 \\
\bottomrule
\end{tabular}
\end{table}

\paragraph{Influence of rollout acceleration on total runtime} Following the theoretical analysis on the complexity, the nesting strategies translate into linear trends on the total runtime. This trend is observed in all the domains and scaled covered (see Figures \ref{fig:gsm8k}, \ref{fig:mbpp}).
The efficiency gain on both metrics increases linearly with the ratio $x\%$ of skipped layers.

\paragraph{Effectiveness of the variance mitigation strategy}

We consider $3$ off-polyciness mitigation strategies, namely \texttt{Base}, \texttt{Uncorrected} and \texttt{Retrace-$\lambda$}.
From Table \ref{tab:strategy_wins_losses_by_skip}, we observe that the \texttt{Uncorrected} strategy leads to the worse win/loss ratio. \texttt{Retrace-$\lambda$} with $\lambda=0.1$ is outperformed by \texttt{Base} and \texttt{Retrace-$\lambda$} with $\lambda=1$. Interestingly, on the most challenging off-policy setting with a skipping ratio $x=0.15$, \texttt{Retrace-$\lambda$} with $\lambda=1$ significantly outperfoms the other variance mitigation approaches. These results on high-dimensional LLM fine-tuning show that Retrace-$\lambda$  can be successfully applied to the class of policy gradient algorithms and to GRPO, extending empirically the algorithmic and applicative scope of the method.

\section{Conclusion}

In this study, we explore the possibility of conducting off-policy RL fine-tuning. 
Specifically, we focus on the goal of achieving more compute efficient rollouts by instantiating the behavior model as nested instances of the target model.
Our main conceptual contribution is to show that it is possible to increase the degree of off-policyness in RL fine-tuning with limited influence on performance. 

\paragraph{Emerging challenges for off-policy roll-outs}

Our controlled experiments show that it is possible to train smoothly, even when the degree of off-policyness increases. 
These results are obtained for fixed size completion generation for the nested behavior model.
However, an increasing number of LLMs can produce adaptive responses, either short for simple problems or long for complex problems. 
The interaction between generating completion off-policy through nesting and its influence on the completion length remains an open research problem.
Furthermore, the nesting strategy may have non uniform interaction effects on the generation length depending on the dataset and model scale.
This opens the door to increasing off-policyness with learnable strategies to handle such interactions effectively. 

\begin{table}[htbp]
\centering
\caption{Strategy Win/Loss Ratio by Skip Ratio. W/L ratio $>1$ indicates more wins than losses. }
\label{tab:strategy_wins_losses_by_skip}
\resizebox{\columnwidth}{!}{%
\begin{tabular}{lcccc}
\toprule
\textbf{Skip} & \textbf{base} & \textbf{uncorrected} & \textbf{retrace\_0.1} & \textbf{retrace\_1.0} \\
\midrule
0.05 & 0.8 {\scriptsize\textcolor{gray}{(4/5)}} & 0.5 {\scriptsize\textcolor{gray}{(1/2)}} & 1.0 {\scriptsize\textcolor{gray}{(3/3)}} & \textbf{2.0} {\scriptsize\textcolor{gray}{(4/2)}} \\
0.1 & 1.0 {\scriptsize\textcolor{gray}{(3/3)}} & 0.3 {\scriptsize\textcolor{gray}{(1/3)}} & \textbf{2.0} {\scriptsize\textcolor{gray}{(6/3)}} & 0.7 {\scriptsize\textcolor{gray}{(2/3)}} \\
0.15 & 2.5 {\scriptsize\textcolor{gray}{(5/2)}} & 0.4 {\scriptsize\textcolor{gray}{(2/5)}} & 0.2 {\scriptsize\textcolor{gray}{(1/4)}} & \textbf{4.0} {\scriptsize\textcolor{gray}{(4/1)}} \\
\midrule
\textbf{Global} & 1.2 {\scriptsize\textcolor{gray}{(12/10)}} & 0.4 {\scriptsize\textcolor{gray}{(4/10)}} & 1.0 {\scriptsize\textcolor{gray}{(10/10)}} & \textbf{1.7} {\scriptsize\textcolor{gray}{(10/6)}} \\
\bottomrule
\end{tabular}%
}
\end{table}

\appendix

\bibliography{references}

\bibliographystyle{icml2026}

\newpage
\appendix
\onecolumn

\section{Experiments details}

\begin{table}[h]
\centering
\small  
\setlength{\tabcolsep}{4pt}  
\begin{tabular}{|l|c|c|c|c|c|}
\hline
\textbf{Model} & \textbf{N} & \textbf{W} & \multicolumn{3}{c|}{\textbf{Skipped layers at ratio $x$}} \\
\cline{4-6}
& & & \textbf{5\%} & \textbf{10\%} & \textbf{15\%} \\
\hline
Qwen2.5-1.5B  & 28  & 1536  & 1 & 3 & 4 \\
Qwen2.5-7B    & 28  & 3584  & 1 & 3 & 4 \\
\hline
\end{tabular}
\caption{Skipped layers for various ratios $x$ on Qwen2.5-Math-Instruct. L = \# of layers, W = hidden layer width.}
\label{tab:skipped_layers}
\end{table}

\paragraph{Training generation details}
Since we are using Instruct models, we consider $E_\text{sft}=0$ epochs for the SFT warm-up stage.
The $\beta$ parameter of GRPO is set to $0.05$. 
To maintain training stability, we apply a symmetric clipping constraint to the importance sampling ratio. We set the clipping parameter $\epsilon = 0.05$, restricting the ratio to the interval $[0.95, 1.05]$. This corresponds to the epsilon argument in the GRPO trainer. 
The batch size is set to $32$ for all models sizes, using gradient accumulation. 
We consider $S=1000$ gradient steps for ReFT, this corresponds to $E_\text{rft}=2.14$, and $E_\text{rft}=43.47$ for GSM8k and MBPP datasets, respectively. 
We use a learning rate of $1e-6$ with a cosine decay. The gradient clipping is set to $1$. The number of iterations to $1$ to avoid noise from other sources of off-policyness in our experiments.
The prompts are formatted using a Qwen-chat template commonly used by practitioners \citep{liu2025understanding}.
For rollouts during training, we set the minimum and maximum length of the generated completions to to $256$ tokens. This implies that all the completions have equal length.  
For training, all the other parameters follow the default from GRPO TRL library \citep{vonwerra2022trl}.

\paragraph{Sampling Parameters and Absolute Continuity.} The theoretical unbiasedness results assume absolute continuity between the reference and current policy distributions. In our implementation using TRL's GRPOTrainer, we employed the default sampling configuration: temperature=1.0, top-p=1.0, and top-k=0. These settings ensure that all tokens in the vocabulary maintain non-zero probability during generation, as the softmax output is not truncated by nucleus (top-p) or top-k filtering. This configuration satisfies the absolute continuity requirement.

\paragraph{Evaluation generation details}
For evaluation, we use a math reasoning benchmark composed of 5 datasets \citep{liu2025understanding}.The temperature is set to $0.6$, the top-p to $0.95$ and the maximum number of tokens to $32$k and a minimum set to $0$ (different setting compared to training). We perform pass@K with $K=1$, implying the model generates $1$ response per problem. This corresponds to a strict setup as the model is only given one single chance to answer correctly. 

\paragraph{Hardware details} The experiments are all conducted on the same hardware setup. We use 4 x H100 NVIDIA GPUs, connected with NVLINK. The models are loaded with DeepSpeed Zero Level 1. The models are loaded with bf16 format. We use KV-cache. The behavior model infers using Hugging Face backend. The GRPO implementation is from TRL library.

\paragraph{Performance metrics and delta to the baseline}
To characterize reasoning performance at test-time, we report the average accuracy on the $5$ math benchmarks \citep{liu2025understanding}. 
To characterize compute efficiency gains at train-time, we report the token speed (total number of tokens processed divided by total runtime), and the total run time (expressed in seconds).
We characterize \texttt{Nested-ReFT} run instances using the relative \textit{delta} ($\Delta$) to the baseline, which is defined for any metric $z$ as $
\Delta(z) = 100 \cdot \frac{(z - z_{\texttt{baseline}})}{ z_{\texttt{baseline}} }
$, where the absolute delta is $ \Delta_{\text{abs} }(z) = (z - z_{\texttt{baseline}}) $.

\end{document}